\documentclass{article}

\usepackage{arxiv}
\usepackage{multirow}
\usepackage{makecell}
\usepackage[most]{tcolorbox}

\usepackage[utf8]{inputenc} % allow utf-8 input
\usepackage[T1]{fontenc}    % use 8-bit T1 fonts      % hyperlinks
\usepackage{url}            % simple URL typesetting
\usepackage{booktabs}       % professional-quality tables
\usepackage{amsfonts}       % blackboard math symbols
\usepackage{nicefrac}       % compact symbols for 1/2, etc.
\usepackage{microtype}      % microtypography
\usepackage{xcolor}         % colors
\usepackage{hyperref} 

\title{Agent-as-a-Service based on Agent Network}

\author{%
  Yuhan Zhu\\
  School of Computer Science\\
  Wuhan University\\
  Wuhan, China \\
  \texttt{zhuyuhan2333@whu.edu.cn} \\
  \And
  Haojie Liu\\
  School of Computer Science\\
  Wuhan University\\
  Wuhan, China \\
  \texttt{ldlornd@whu.edu.cn} \\
  \And
  Jian Wang\thanks{Corresponding author. Email: \texttt{jianwang@whu.edu.cn}}\\
  School of Computer Science\\
  Wuhan University\\
  Wuhan, China \\
  \texttt{jianwang@whu.edu.cn} \\
  \And
  Bing Li\\
  School of Computer Science\\
  Wuhan University\\
  Wuhan, China \\
  \texttt{bingli@whu.edu.cn} \\
  \And
  Zikang Yin\\
  School of Computer Science\\
  Wuhan University\\
  Wuhan, China \\
  \texttt{2020302111233@whu.edu.cn} \\
  \And
  Yefei Liao\\
  School of Computer Science\\
  Wuhan University\\
  Wuhan, China \\
  \texttt{1206232012@qq.com} \\
}

\begin{document}

\maketitle

\begin{abstract}
The rise of large model-based AI agents has spurred interest in Multi-Agent Systems (MAS) for their capabilities in decision-making, collaboration, and adaptability. While the Model Context Protocol (MCP) addresses tool invocation and data exchange challenges via a unified protocol, it lacks support for organizing agent-level collaboration. To bridge this gap, we propose Agent-as-a-Service based on Agent Network (AaaS-AN), a service-oriented paradigm grounded in the Role-Goal-Process-Service (RGPS) standard. AaaS-AN unifies the entire agent lifecycle, including construction, integration, interoperability, and networked collaboration, through two core components:
(1) a dynamic Agent Network, which models agents and agent groups as vertexes that self-organize within the network based on task and role dependencies;
(2) service-oriented agents, incorporating service discovery, registration, and interoperability protocols. These are orchestrated by a Service Scheduler, which leverages an Execution Graph to enable distributed coordination, context tracking, and runtime task management. We validate AaaS-AN on mathematical reasoning and application-level code generation tasks, which outperforms state-of-the-art baselines. Notably, we constructed a MAS based on AaaS-AN containing agent groups, Robotic Process Automation (RPA) workflows, and MCP servers over 100 agent services. We also release a dataset containing 10,000 long-horizon multi-agent workflows to facilitate future research on long-chain collaboration in MAS.
  % The abstract paragraph should be indented \nicefrac{1}{2}~inch (3~picas) on
  % both the left- and right-hand margins. Use 10~point type, with a vertical
  % spacing (leading) of 11~points.  The word \textbf{Abstract} must be centered,
  % bold, and in point size 12. Two line spaces precede the abstract. The abstract
  % must be limited to one paragraph.
\end{abstract}

\section{Introduction}
% 随着基于大模型的AI AGent的发展，多智能体系统（Multi-Agent System, MAS）在自动化决策、任务协同和复杂环境适应性方面展现出广泛的应用前景。与此同时，云计算、边缘计算和服务计算等新型计算模式与环境的快速演变推动着软件系统向网络化、服务化与智能化等方向演进。 然而，现有的多智能体系统主要由智能体工作流来组织多智能体协作，缺乏能够从智能体构造、智能体集成、互操作和多智能体协作的全流程自动化方法。
With the rise of large model-based AI agents, Multi-Agent Systems (MAS) are gaining traction for their potential in automated decision-making, collaborative task execution, and adaptability to complex environments~\cite{DBLP:journals/corr/abs-2503-13415}. Meanwhile, the rapid evolution of computing paradigms—such as cloud, edge, and service-oriented computing—is driving software systems toward greater connectivity, intelligence, and service modularity~\cite{DBLP:journals/corr/abs-2402-01968}. However, current MAS implementations still largely rely on agent workflows to coordinate collaboration, lacking end-to-end automation across agent construction, integration, interoperability, and collaboration.\par
% 为了解决智能体工具调用复杂、数据孤岛和数据交互格式不统一的问题，模型上下文协议（MCP）为大型语言模型（LLM）应用提供统一的通信接口，支持实时、双向的上下文信息交换，从而显著提升模型的可扩展性和响应相关性，便于大量的工具集成成为智能体。然而mcp协议并没有规定智能体之间的协作范式，当智能体数量激增后，如何有效地组织多智能体实现用户需求成为了更加迫切的问题。
To address challenges such as complex tool invocation, data silos, and inconsistent data exchange formats, the Model Context Protocol (MCP) \cite{DBLP:journals/corr/abs-2503-23278} provides a unified communication interface for Large Language Model (LLM) applications. It supports real-time context exchange, significantly improving model scalability and response relevance, and enabling seamless integration of a wide range of tools into agents. However, while MCP facilitates tool integration, it does not define collaboration paradigms among agents. As the number of agents rapidly increases, effectively organizing multi-agent systems to fulfill user needs has become a more pressing challenge.
% Agent2Agent进一步为智能体之间的协作制定了范式，能够支持不同的智能体框架接入A2A协议实现多智能体协同，并且提供了任务状态管理、智能体能力发现和数据安全的保障。然而如何针对用户需求自动发现智能体，并组织多智能体进行有效地协同，使得智能体成为开箱即用的服务，并支持智能体系统持续扩展，与现有的软件系统实现无缝衔接，依然缺乏服务化的互操作标准。
Agent2Agent (A2A)~\cite{hou2025modelcontextprotocolmcp} further establishes a collaboration paradigm for agents, enabling multi-agent collaboration across heterogeneous agent frameworks through the A2A protocol. It also provides support for task state management, agent capability discovery, and data security. However, challenges remain in automatically discovering appropriate agents based on user needs and organizing them for effective collaboration. Making agents truly plug-and-play services that support scalable multi-agent systems and seamlessly integrate with existing software infrastructures still lacks standardized, service-oriented interoperability protocols. \par
% In this paper, 我们提出Agent as a Service based on Agent Network(AaaS-AN), 以RGPS标准为基础，旨在以服务化智能体范式来统一包含智能体构造、集成、互操作和网络化协作的智能体生命周期全过程，实现多智能体系统的开箱即用。AaaS-AN包含智能体网络、服务化智能体、调度器三大板块。智能体网络提供了智能体构造和网络化协同的基础，其中以智能体和智能体组为节点，动态路由为边，形成分布式自组织的智能体网络。服务化智能体以智能体组为基本单元，包含分布式智能体服务的运行、服务注册与发现和智能体通信协议。调度器负责驱动智能体服务协同运行，提出执行图作为运行时协议来解决分布式智能体服务中涉及到的上下文管理、多智能体协作、任务状态维护和智能体系统运维。
% 我们在数学推理和应用级代码生成问题上探究了基于 AaaS-AN 构建的多智能体系统的能力。 我们进一步集成了包含服务化智能体、RPA流程和MCP Server的超过200个智能体服务所构成的多智能体系统，公开了包含10000个长链路多智能体协同的数据集，探究了该数据集中反映出的关于多智能体系统在长链路流程中的问题和潜在的解决方案，以帮助多智能体系统在长链路流程上的进一步研究。
In this paper, we propose Agent-as-a-Service based on Agent Network (AaaS-AN), a service-oriented agent paradigm built upon the RGPS standard \cite{DBLP:books/sp/wsf14/WangFZHHZ14}. AaaS-AN aims to unify the entire lifecycle of agents—including construction, integration, interoperability, and networked collaboration—under a service-based framework, enabling plug-and-play multi-agent systems. AaaS-AN consists of two core components: the Agent Network and service-oriented agents, collaborated by the Service Scheduler using interoperability protocols.
The Agent Network provides the foundation for agent construction and networked collaboration. It models agents and agent groups as vertexes, and dynamic routing as edges, forming a distributed, self-organizing network. The Service-Oriented Agent component treats agent groups as the basic units and supports distributed agent service execution, service registration and discovery, and standardized agent communication protocols. The Service Scheduler coordinates agent service execution and introduces an Execution Graph as a runtime protocol to handle context management, multi-agent collaboration, task state tracking, and overall system operations in distributed multi-agent systems.

We evaluate the capabilities of AaaS-AN-based multi-agent systems in mathematical reasoning and application-level code generation tasks. Furthermore, we integrate a large-scale multi-agent system composed of over 100 agent services, including agent group services, RPA workflows, and MCP Servers. We also release a dataset containing 10,000 long-chain multi-agent collaboration flows, and investigate the challenges and potential solutions revealed within these long-chain flows, contributing to future research on long-horizon multi-agent collaboration.
% 我们的核心贡献点总结如下：
% 1. 我们提出了Aaas-AN，a service-oriented agent paradigm to make each agent running as a vertex in the Agent Network and form agent services to collaborate to better solve the user requirements.
% 2. 我们设计了包含服务注册与发现、服务通信协议和服务调度器，实现了服务化智能体的开箱即用，让智能体以服务的形式直接触达用户，在数学推理和应用级代码生成这两类任务上验证了AaaS-AN的效果。
% 3. 我们集成了包含智能体组服务、RPA流程和MCP server等100+智能体服务，并公开了包含10000个长链路的多智能体协同数据集，为长链路流程的多智能体协同提供了研究基础。

Our main contributions are summarized as follows:
\begin{itemize}
    \item We propose AaaS-AN, a service-oriented agent paradigm that enables each agent to operate as a vertex within an Agent Network and collaborate as agent services to more effectively fulfill user requirements.
    \item We design and implement a plug-and-play framework for service-oriented agents, including service registration and discovery, interoperability protocols, and the service scheduler. This allows agents to be accessed directly as services by users. AaaS-AN outperforms baselines on mathematical reasoning and application-level code generation tasks.
    \item We integrate a large-scale multi-agent system consisting of over 100 agent services—including agent groups, RPA workflows, and MCP Servers. We release a dataset of 10,000 long-chain multi-agent collaboration flows. This provides a foundation for research on multi-agent collaboration in long-chain flows.
\end{itemize}
\section{Related Work}
\subsection{Multi-agent System}
% 基于大语言模型的智能体，在网页导航任务中，Gur等人[7]提出WebAgent用于网页自动导航，其使用预训练模型提取HTML信息，再结合代码生成模型生成操作代码，帮助用户实现需求到网页操作的自动化。Park等人[2]构建出“斯坦福小镇”这一交互式沙盒环境研究基于大语言模型的实例化后的25个代理是否能够实现可信的人类行为模拟，研究得出代理架构中以观察、规划和反思为关键组件各自对代理行为的可信性起到了至关重要的作用。 
In the context of web navigation tasks, language model-based agents have been increasingly utilized. Gur et al.\cite{gur2023real} introduced WebAgent, which automates web navigation by utilizing a pre-trained model to extract HTML information. This is then combined with a code generation model to produce the necessary operation code, assisting users in automating the process from requirements to web actions. Additionally, Park et al. \cite{park2023generative} developed an interactive sandbox environment called "Stanford Town" to study whether 25 instantiated agents, based on large language models, could reliably simulate human behavior. Their research concluded that the key components of the agent architecture—observation, planning, and reflection—are crucial in determining the credibility of the agent's behavior.

% 在具身智能领域，为了解决数据稀缺问题，Wang等人进一步将“斯坦福小镇”拓展至了真实环境中，以机器人为载体，解决了由于收集真实世界数据的成本过高导致的在具身智能领域探索尺度定律的重重困难，实现了从仿真到真实的一步。Liu等人提出了基于对话的音频-视觉具身智能导航框架CAVEN，智能体通过与人类或预设神谕（oracle）互动来解决导航任务。
In the field of embodied intelligence, to address the issue of data scarcity, Wang et al. \cite{wang2024grutopia} further expanded the "Stanford Town" into real-world environments, utilizing robots as carriers. This expansion overcame the challenges posed by the high cost of collecting real-world data, which had previously hindered the exploration of scaling laws in embodied intelligence. This development marked a significant step from simulation to reality. Additionally, Liu et al. \cite{liu2024caven} proposed a dialogue-based audio-visual embodied intelligence navigation framework, CAVEN, where agents solve navigation tasks through interactions with humans or a pre-set oracle. 

% 在软件工程领域，Qian等人提出的多智能体的链式交互框架ChatDev，探索了从需求分析到应用级别代码生成的全自动化过程，有效节省大量的人力成本，在软件工程全流程自动化的复杂任务上具有一定的解决能力。由于智能体框架的多样性和多智能协作的复杂性，Chen等人提出智能体互联网，首次引入智能体注册和发现来组织复杂异构的多智能体，并采用状态机来控制多智能体的协作，开始借助服务化的方式实现多智能体系统。Wang等人等探讨了利用LLMs进行推荐系统中的用户行为模拟的潜力，通过设计用户画像、记忆和行为模块，使得每个智能体可以与推荐系统通过一对一聊天或一对多社交广播进行交互。AlpacaFarm框架研究了一种基于LLMs的人类反馈模拟器，其成本比众包人工反馈低45倍，同时研究者发现基于奖励学习的方法可以显著优于监督微调（Supervised Fine-Tuning，SFT），与人类在反馈中不断学习具有高度一致性。 AgentsPLC方法通过多智能体网络实现PLC代码的自动生成。MetaGPT则使用共享信息池的通信结构，智能体以发布和订阅的方式与共享消息池进行互动，获取自己所需的信息，并发布其他智能体需要的信息，推进任务流程。
In the field of software engineering, Qian et al.\cite{qian2023communicative} introduced the multi-agent chain interaction framework ChatDev, which explores the fully automated process from requirements analysis to application-level code generation. This approach effectively saves significant manpower costs and demonstrates certain capabilities in solving complex tasks in the full automation of software engineering workflows. Due to the diversity of agent frameworks and the complexity of multi-agent collaboration, Chen et al.\cite{chen2024internet} proposed the Agent Internet, which introduces agent registration and discovery for organizing complex and heterogeneous multi-agent systems. They employed state machines to control multi-agent collaboration, marking the initial steps toward realizing multi-agent systems through service-oriented approaches. Wang et al. \cite{wang2025user}explored the potential of using LLMs for user behavior simulation in recommendation systems. By designing user profiles, memory, and behavior modules, each agent can interact with the recommendation system via one-on-one chats or one-to-many social broadcasts. The AlpacaFarm framework \cite{dubois2023alpacafarm} studied a human feedback simulator based on LLMs, which is 45 times cheaper than crowdsourced human feedback. The researchers also found that reward-learning-based methods significantly outperformed supervised fine-tuning (SFT) and showed a high degree of consistency with human feedback in the learning process. The AgentsPLC\cite{liu2024agents4plc} method achieves automatic generation of PLC code through a multi-agent network. MetaGPT\cite{hong2023metagpt} uses a shared information pool communication structure, where agents interact with the shared message pool in a publish-subscribe manner. 

% 然而现有的多智能体系统通信内容还是以自然语言文本为主，无法满足服务部署的结构化需求；智能体所依赖的工具链（如API、服务接口）和智能体间的交互需要采用服务化方式进一步规范和统一，以服务的形式落地具体应用场景，与现有软硬件系统无缝衔接；对于用户与多智能体系统，智能体需要依赖服务（明确定义的功能的集合）来实现与人的交互。因此，如何高效、稳定地传递结构化信息，赋能智能体网络准确完成用户需求，将会是服务化智能体系统的研究重点。
However, existing multi-agent systems primarily rely on natural language text for communication, which fails to meet the structured requirements of service deployment. The toolchains that agents depend on (e.g., APIs, service interfaces) and the interactions between agents need to be further standardized and unified through a service-oriented approach. This will enable the practical application of these systems in specific scenarios and ensure seamless integration with existing software and hardware systems. For user interactions with multi-agent systems, agents must rely on services (a clearly defined set of functions) to engage with humans. Therefore, efficiently and reliably transmitting structured information to empower agent networks to accurately fulfill user requirements will be a key research focus in service-oriented agent systems.

\subsection{Agent Paradigm}
The architecture of intelligent agents encompasses several key components: context-aware memory~\cite{park2023generative}, planning~\cite{wang2023plan}, role-playing~\cite{wang2023unleashing}, and tool utilization~\cite{qin2023toolllm}. Well-designed architectures significantly enhance agent capabilities. Chain of Thought~\cite{wei2022chain} (CoT) guides LLMs in "stepwise reasoning", decomposing complex tasks into smaller subtasks to improve reasoning performance. An enhanced CoT technique, self-consistency~\cite{wang2022self}, replaces simple greedy decoding by first sampling a diverse set of reasoning paths rather than selecting the most straightforward route, then selecting the most consistent answer through marginalizing these sampled reasoning paths. The ReAct~\cite{yao2023react} framework integrates reasoning with action, expanding LLMs' action space to enable natural language-based reasoning generation and environmental interaction. Reflexion~\cite{shinn2023reflexion} incorporates reinforcement learning mechanisms to endow agents with dynamic memory and self-reflective capabilities, thereby improving decision-making efficiency and reasoning performance. HuggingGPT~\cite{shen2023hugginggpt} framework employs LLMs as task planners, automatically selecting optimal models from HuggingFace platforms based on task requirements and generating final responses through execution results, substantially enhancing LLMs' scalability and adaptability in complex tasks. When conducting self-evaluation, LLMs often exhibit excessive confidence or high randomness, generating stubborn or inconsistent feedback that undermines reflection effectiveness. To address this limitation, the Self-Contrast~\cite{selfself} method provides LLMs with multiple perspectives to mitigate stubborn biases.

\section{AaaS-AN}
\subsection{Overview}
% 基于大语言模型的智能体系统往往面临单一智能体能力受限、难以完全满足复杂任务需求的问题，在利用多智能体协同工作时，又因边界模糊而难以高效协作实现任务目标。 我们基于Role-Goal-Process-Service (RGPS) 设计了AaaS-AN，分为智能体网络和服务化智能体两部分。智能体网络以领域知识为基础建模智能体角色，围绕目标建模智能体组，以智能体路由实现复杂流程，通过网络化方式组织多智能体协作。Agent as a Service将智能体组作为基本单元将智能体服务化，执行任务时，基于调度器与智能体服务的互操作协议实现智能体服务的调度，同时将执行图作为运行时协议存储和隔离不同任务的结构化上下文。基于AaaS-AN可以实现多智能体系统包含智能体构造、集成、运行和协作的全流程自动化，如图所示。
LLM-based agent systems often struggle with the limitations of single-agent capabilities, making it difficult to fully address complex task requirements. While multi-agent collaboration offers a potential solution, vague role boundaries frequently hinder effective cooperation and task execution.

\begin{figure*}[t]
\centering
\includegraphics[width=\linewidth]{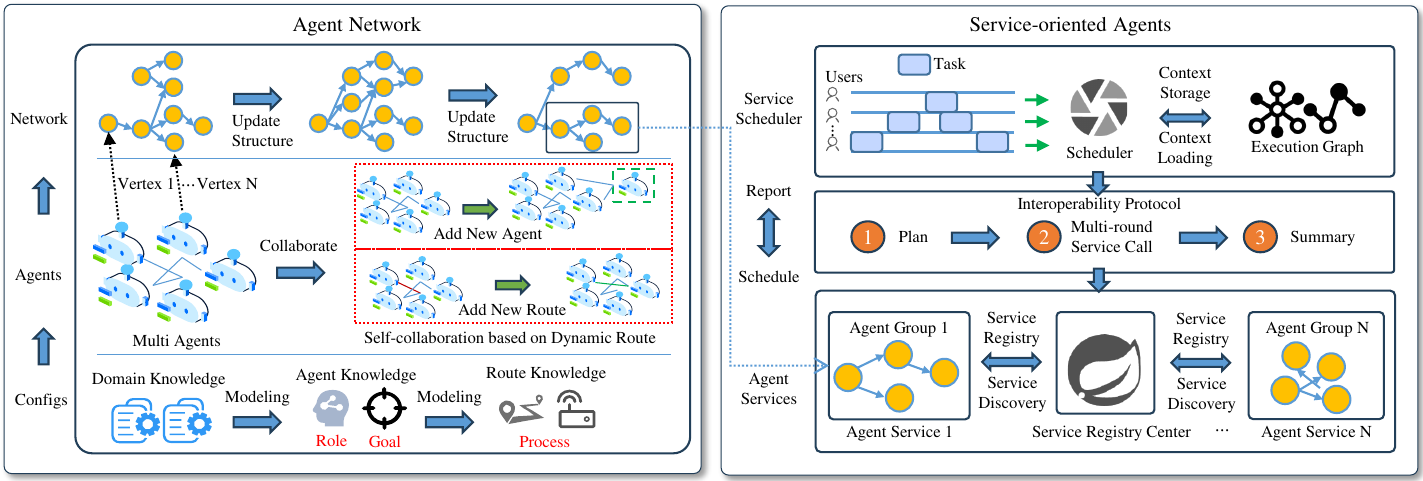}
\caption{The overview framework of AaaS-AN.} \label{1.Overview.pdf}
\end{figure*}

To address these challenges, we propose AaaS-AN, a Role-Goal-Process-Service (RGPS)-driven architecture composed of two main components: the Agent Network and service-oriented agents.
The Agent Network models agent roles based on domain knowledge, forms agent groups centered around specific goals, and employs agent routes to support complex execution flows. This network-based organization enables structured and scalable multi-agent collaboration.
Service-oriented agents encapsulates agent groups as service units. During task execution, a scheduler coordinates these services via the interoperation protocol, while a execution graph is maintained to store and isolate structured contexts across tasks.
Built upon AaaS-AN, it supports full-process automation for multi-agent systems, including agent construction, integration, execution, and collaboration, as illustrated in Figure \ref{1.Overview.pdf}.

\subsection{Agent Network}
% 智能体网络由大量的智能体和智能体组作为节点，路由作为边来构成。网络中的任意节点都能被其余节点通过路由进行访问。一个智能体组内包含多个智能体，而智能体组又能作为网络中的抽象节点被其他节点通过路由访问，形成递归形式的调用。智能体网络中的任意节点都能接收用户任务，并沿着网络执行得到最终汇总输出。同一时刻智能体网络能够并行处理多个用户任务，每个任务拥有相互隔离的上下文，但同时节点能够根据策略共享经过该节点的任务上下文。整个网络中的所有节点及节点间的边均可以动态变化和调整。
The agent network is dynamically structured, where both individual agents and agent groups serve as vertexes, and routes form the edges. Each vertex is accessible via routes, enabling decentralized communication. Agent groups encapsulate multiple agents and can be treated as abstract vertexes, supporting recursive invocation across the network.
Any vertex can receive user tasks, triggering distributed execution and yielding aggregated outputs. The network supports concurrent task execution with isolated contexts, while allowing for context sharing at specific vertexes under configurable policies. Both the topology of vertexes and the routing edges are dynamically adjustable, enabling adaptive reconfiguration based on runtime requirements.
% 智能体角色知识化
\subsubsection{Agent Role}
% 在RGPS需求元模型中，角色作为在理解需求角色之后对单个智能体进行建模的知识指导，所形成的单个智能体知识(A)主要包含角色名称(A^n)、角色描述(A^d)、系统提示词(A^p)、输入参数(A^i)、输出参数(A^o)、智能体逻辑编码(A^c)六个部分。智能体知识表示如下：
In the RGPS requirements meta-model, a role serves as the knowledge foundation for modeling individual agents in the agent network. The resulting agent knowledge (denoted as $A$) comprises six key components:
\begin{equation}
A=\{A^n,A^d,A^p,A^i,A^o,A^c\}    
\end{equation}
where $A^n$ denotes the name of the agent role, $A^d$ denotes the description, $A^p$ denotes the system prompt, $A^i$ and $A^o$ denotes the structured input and output parameters, and $A^c$ denotes the logic code.
% 角色名称概括了智能体的核心功能，例如在代码生成任务中，存在一个专门编写代码的程序员智能体。角色描述以文本的形式描述了智能体的功能职责。系统提示词作为该智能体的输入给大语言模型用于提升智能体角色扮演和任务执行前后的一致性。输入参数和输出参数作为智能体的细节描述和参数约束，一方面定义智能体所需要的数据源和所能提供的数据，另一方面作为运行时的校验，增强多智能体协同时上下游智能体的因果关系。智能体逻辑编码用于将输入参数转化为输出参数，构成完整的智能体。
The role name summarizes the core functionality of an agent, for example, in a code generation task, there may be a "programmer" agent specifically responsible for writing code.
The role description provides a textual explanation of the agent’s responsibilities and functional scope.

The system prompt is input to the large language model, enhancing the alignment of the agent and maintaining consistency across the task lifecycle.
The structured input and output parameters serve as both detailed descriptions and operational constraints for the agent, which define the context the agent requires and produces, and also function as runtime checks that reinforce causal relationships between upstream and downstream agents during multi-agent collaboration, as well as providing valuable context to chat with LLMs.

Finally, the agent logic code transforms input parameters into output parameters, completing the functional definition of the agent. \par
\subsubsection{Agent Group}
Agent Group is the set of goal-oriented agents for better collaboration. Based on goal modeling in the RGPS meta-model, the construction of goal-oriented agent groups relies on explicit goal decomposition and collaboration mechanisms. Within a group, each agent assumes a specific role, and the system-level objective is achieved through a structured goal decomposition strategy.
\paragraph{Goal Decomposition:}
High-level global objectives are decomposed into multiple sub-goals, and agent roles are assigned or organized around each sub-goal.
\paragraph{Role Assignment and Group Organization:}
Based on the nature of each sub-goal and the knowledge representation of agent roles, agents are assembled into a group such that each agent can effectively perform its designated task. The agent group is centered around a system-level goal, and its structured knowledge (denoted as $G$) consists of the following components:
\begin{equation}    
G=\{G^n,G^d,G^p,G^i,G^o,G^A\}
\end{equation}
where $G^n$ denotes the name of the group, $G^d$ denotes the goal description, $G^p$ denotes the system prompt of the group, $G^i$ and $G^o$ denotes the structured input and output parameters to align task context with the goal of the group, and $G^A$ denotes the agents within the group.
The agent group name summarizes the core functionality associated with a system-level goal.
The goal description is a textual summary of the overall responsibilities and intended objectives of the group.
The group prompt is provided to the language model as input to enhance consistency in task execution and collaboration among all agents within the group.
The input and output parameters define the data dependencies and serve as operational constraints at the group level, allowing the agent group, as a knowledge unit, to collaborate further with other agent groups or individual agents.
In code generation tasks, there may be a dedicated agent group responsible for analyzing requirements based on user tasks, including a requirement refinement agent and a programming language selection agent, each responsible for translating high-level requirements into executable development tasks.
\subsubsection{Agent Route}
% 智能体路由是为了能够将多智能体协作流程进行知识化建模，通过信息交换和决策协调，使得智能体网络中的节点能够有效协同，智能体网络中包含多种不同类型的路由，通过路由实现组内智能体的自主协同和跨组智能体的自主协同。
The agent route is designed to enable knowledge-based modeling of multi-agent collaborative processes, based on "Process" in RGPS. By facilitating context exchange and coordinated decision-making, it allows vertexes within the agent network to collaborate effectively. The network supports multiple types of routes, including "HARD", "SOFT", and "EXT", enabling both intra-group and inter-group self-collaboration among agents.

% 硬路由是为了确定智能体网络中的协作结构而设计的。一方面，某些智能体之间需要固定的协作模式来完成任务，可以通过引入人为的先验知识设计硬路由将工作流中的固定流程作为知识引入到智能体网络中。另一方面，在智能体网络的动态变化过程中，大量的任务执行会生成相应的执行轨迹，通过执行轨迹与任务执行的相关性能够确定一些固定的网络结构，这些结构借助硬路由的形式得以保留。
\paragraph{HARD Route:}Hard route is designed to define and preserve the structural collaboration patterns within the agent network. On the one hand, certain agents require predefined interaction sequences to accomplish specific tasks. These can be encoded as hard routes using prior domain knowledge, injecting fixed workflows as structured knowledge into the network. On the other hand, during the dynamic evolution of the agent network, a large number of execution trajectories are generated. By analyzing the correlation between execution traces and task success, recurring structural patterns can be identified and retained through hard routes, enabling the network to accumulate and reuse effective collaboration structures.

% 软路由是为了智能体组内的自主协作结构而设计的。 在同一个智能体组内，智能体组节点接收到组级别任务，根据不同的任务，需要通过软路由组织智能体组内的所有智能体节点进行协同。
\paragraph{SOFT Route:}Soft route is designed to support self-collaboration structures within an agent group. When an agent group vertex receives a group-level task, it must dynamically organize the constituent agents based on task-specific requirements. Soft route enables this collaboration by organizing intra-group agents in a flexible and adaptive manner, facilitating effective collaboration aligned with the task.

% 扩展路由是为了跨智能体组的自主协作结构而设计的。同一个智能体组在执行任务时并不一定能完全与任务匹配，当任务上下文不足以支撑该智能体组顺利完成任务时，需要智能体组节点通过扩展路由寻找智能体网络中其他节点自主协同以实现任务目标。
\paragraph{EXT Route:}Extended route is designed to support self-collaboration across agent groups. During task execution, an individual agent group may not always possess sufficient capabilities or context to fulfill the task independently. In such cases, extended route enables the agent group vertex to proactively discover and collaborate with other vertexes in the agent network. It allows the system to dynamically expand its collaboration scope beyond the original group, facilitating task completion.\par
% 基于上述三种路由能够有效地将具备先验知识的固定工作流和需要动态自主协同流程有效结合起来，实现智能体网络中节点的有效协作，以最终更好地实现用户目标。
Based on the above three types of route mechanisms, it is possible to effectively integrate fixed workflows enriched with prior knowledge and dynamically coordinated agent interactions. This integration facilitates efficient collaboration among vertexes in the agent network, ultimately enabling more effective achievement of user goals.
% \subsection{Service-Oriented Agents}
% 123
% \subsubsection{Agent Service}
% \subsubsection{Service Scheduler}
% \subsubsection{Interoperability Protocol}
% \subsubsection{Execution Graph}
% 123

\section{Experiments}

%在本章，我们分别对 AaaS-AN 在数学推理、应用级代码生成和现实世界长链路工作流任务中的表现进行了测试。
In this section, we conducted a systematic evaluation of AaaS-AN's performance across three distinct domains: mathematical reasoning, application-level code generation, and real-world long-chain workflow tasks. The framework's capabilities were rigorously examined through quantitative benchmarking, qualitative case studies, and comparative analysis against state-of-the-art baselines in each respective task category.

\subsection{Mathematical Reasoning}

%为了评测多智能体框架在数学推理任务上的泛化性，我们在 MATH 数据集的七类问题中分别选取了 72 个问题，总计 504 个问题，用于测试 AaaS-AN 和其他基线的表现。
For comprehensive evaluation of our multi-agent framework's generalization capacity on mathematical reasoning tasks, we constructed a balanced test suite comprising 72 problems per category (504 total) from the MATH benchmark. This stratified sampling approach enables systematic analysis of performance variations across mathematical domains.

%对于基线方法，我们则选取了当下最先进的多智能体框架 MetaGPT、AutoGen 和 MACM。其中 MetaGPT 和 AutoGen 是当下非常热门的多智能体框架，而 MACM 则提出了一种与众不同的数学推理方法，并声称在 MATH 数据集中获得了提升。
\paragraph{Experimental Setups:} We benchmark against three state-of-the-art multi-agent frameworks:
\begin{itemize}
    \item MetaGPT: The first framework to introduce workflows through meta-programming, enabling agents with human-like domain expertise to verify intermediate results and reduce errors.
    \item AutoGen: A popular multi-agent framework proposed by Microsoft, capable of decomposing and resolving complex tasks via multi-round dialogues. The framework demonstrates strong performance and generalization capabilities across multiple domains.
    \item MACM: An advanced multi-agent framework for solving complex mathematical problems, which uses condition mining to solve mathematical problems, expected to improve the reasoning capabilities of large language models on advanced mathematical problems.
\end{itemize}

%在对问题的结果进行测试时，为了提升评测结果的准确性，我们抛弃了精确匹配的方式，而使用当下最先进的大语言模型进行评测智能体输出答案与标准答案的语义一致性，评测 prompt 如下：
To enhance assessment reliability, we use state-of-art LLM for semantic consistency scoring between agent outputs and reference solutions instead of exact matching, using the prompt template below:

\begin{tcolorbox}
    system: 
    
    You are an experienced mathematics teacher with a strong grasp of logical reasoning and precise calculations, capable of quickly identifying the core of mathematical problems and evaluating the consistency between answers and solution processes.
    
    User: 
    
    Here is the math problem: \{problem\}
    with standard solution:
    \{solution\}
    The student's answer is:
    \{answer\}
    Please check whether the answer is correct or not. 
    Please answer in True or False directly without any 
    additional explanations.
\end{tcolorbox}

\begin{table}[htbp]
    \centering
    \begin{tabular}{ccccc}
        \hline
         Model &  Method & Accuracy & Token Cost & Time(s) \\
         \hline
         \multirow{4}{*}{qwen2.5-32b-instruct} & MACM & 35.13\% & 4429.05 & 44.26 \\
         \multirow{4}{*}{} & MetaGPT & 57.52\% & \textbf{2264.17} & \textbf{37.91} \\
         \multirow{4}{*}{} & AutoGen & \underline{57.85\%} & 5688.95  & \underline{40.38} \\
         \multirow{4}{*}{} & AaaS-AN & \textbf{63.62\%} & \underline{2297.42} & 41.77 \\
         \hline
         % \multirow{4}{*}{gpt-4o-mini} & IO & 52.21\% & 951.73 & 18.42s \\
         % \cline{2-5}
         % \multirow{4}{*}{} & MACM & 32.94\% & 4520.60 & 43.61s \\
         % \cline{2-5}
         % \multirow{4}{*}{} & AutoGen & 55.16\% & 5981.19  & 41.64s \\
         % \cline{2-5}
         % \multirow{4}{*}{} & AaaS-AN & 50.40\% & 2277.57 & 40.31s \\
         % \Xhline{1.5pt}
    \end{tabular}
    \caption{The performance on the mathematical reasoning tasks of AaaS-AN and other baselines. The top scores are in bold, with the second-highest underlined in Quality. Given that the token consumption and time consumption are comparable to those of other state-of-the-art models, AaaS-AN has significantly improved the accuracy(5.77\%).}
    \label{The performance on the mathematical reasoning tasks}
\end{table}

\paragraph{Results and Analysis:}
%实验结果如表 label 所示。在数学推理任务中，AaaS-AN 的准确率显著高于其他基线模型，平均高出 5.77%。与此同时，AaaS-AN 在解决问题时的 Token 消耗和时间消耗与现行最优方法十分接近。在对随机取样的 504 个问题进行分析时，AaaS-AN 的平均 Token 消耗仅比最优方法多 1.47%，平均时间多 10.18%。鉴于调用大语言模型 API 时存在网络不稳定因素，我们认为 10% 的时间偏差属于合理范围。
The experimental results are shown in Table label. In the mathematical reasoning task, AaaS-AN achieved a significantly higher accuracy rate than other baseline models, with an average improvement of 5.77\%. Meanwhile, the Token consumption and time consumption of AaaS-AN when solving problems were very close to those of the current state-of-the-art methods. In an analysis of 504 randomly sampled problems, the average Token consumption of AaaS-AN was only 1.47\% higher than that of the best method, and the average time was 10.18\% longer. Considering the network instability that exists when calling the large language model API, we believe that a 10\% deviation in time is within a reasonable range.

%与现有的主流多智能体框架相比，得益于结构化上下文的优势，AaaS-AN 中的智能体信息传递，更加高效和精准。因此，AaaS-AN 能够保持出色的问题解决效率，并提高问题解决的成功率。与此同时，MACM 的低准确率可能是因为其多智能体协作方式对某一类问题进行了特化。而我们的数据集更加强调问题的多样性，为了用一个框架解决这些数学问题，多智能体框架需要具备更好的泛化性。这正是 AaaS-AN 的优势所在。
Compared with the mainstream multi-agent frameworks currently in use, the information transmission among agents in AaaS-AN is more efficient and precise, thanks to the advantage of structured context. As a result, AaaS-AN is able to maintain excellent problem-solving efficiency and enhance the success rate of problem resolution. Meanwhile, the low accuracy rate of MACM may be attributed to the specialization of its multi-agent collaboration approach for a certain type of problem. In our dataset, which emphasizes problem diversity, a multi-agent framework aiming to solve these mathematical problems needs to possess better generalizability. This is where AaaS-AN excels.

\subsection{Application-level Code Generation}
% 为了评估AaaS-AN在0样本复杂生成任务中的能力，我们选择在应用级代码生成任务上的SRDD和ProgramDev两个benchmark上展开实验。 SRDD由ChatDev提出，it comprises 1,200 software task prompts that have been carefully categorized into 5 main areas: Education, Work, Life, Game, and Creation. All these areas are further divided into 40 subcategories, and each subcategory contains 30 unique task prompts. ProgramDev则专注于发现多智能体协作编码中可能忽略的深层次异常导致应用级代码生成效果不佳，这个 benchmark 汇集了30个以经典游戏为主的轻量级程序，覆盖多样化的交互逻辑与功能实现。
To evaluate the capability of AaaS-AN in zero-shot complex generation tasks, we conduct experiments on two benchmarks focused on application-level code generation: SRDD and ProgramDev.
SRDD, proposed by ChatDev, comprises 1,200 software task prompts that are meticulously categorized into five main domains: Education, Work, Life, Game, and Creation. Each domain is further divided into 40 subcategories, with each subcategory containing 30 unique task prompts.
ProgramDev, on the other hand, targets the discovery of deep, often-overlooked failures in multi-agent collaborative coding that may degrade application-level code generation performance. This benchmark includes 30 lightweight programs inspired by classic games, covering a diverse range of interaction logic and functional implementations.
\paragraph{Experimental Setups:}We benchmark AaaS-AN against several state-of-the-art multi-agent systems as well as leading large language models used as single-agent systems.
\begin{itemize}
    \item ChatDev\cite{DBLP:conf/acl/QianLLCDL0CSCXL24}: A chat-powered software development framework that integrates LLM-based agents with diverse social roles, enabling them to autonomously collaborate and generate comprehensive solutions through multi-agent interaction.
    \item GPTSwarm\cite{DBLP:conf/icml/ZhugeWKFKS24}: A LLM-based agents as computational graphs, it implements operations as nodes and information flow between operations as edges to solve the problems by recursively combined graphs.
\end{itemize}
We adopted Quality, Token, Cost, and Time to measure the performances. Quality is defined as the product of task success rate, code completion rate, and executability, providing a comprehensive measure of the accuracy, completeness, and usability of multi-agent systems in application-level code generation tasks. 
AaaS-AN and all baseline methods are instantiated following the software agent team structure used in ChatDev, including agents for requirement analysis, coding, code review, testing, and documentation editing, in order to evaluate their performance.
\paragraph{Results and Analysis:}
% AaaS-AN在SRDD和ProgramDev两个benchmark上都呈现出了更为优秀的性能。
% 与多智能体框架Chatdev和GPTSwarm相比，AaaS-AN在不同的基础大模型上都有不同程度的提升。AaaS-AN较Chatdev能够显著减小token消耗，因为AaaS-AN减少了不必要的历史消息和Chat过程，特别是在代码生成任务中频繁的生成的差异性不大但篇幅冗长的代码，大模型很难从长上下文中的微小差异里捕捉到代码问题。AaaS-AN利用其自主协同机制主动在发现代码无变更时主动进行反思，从而矫正了大量无效的代码生成过程，以提升性能的同时减少无效的智能体执行所带来的token消耗。
AaaS-AN exhibits consistently superior performance on both the SRDD and ProgramDev benchmarks.
In comparison with state-of-the-art multi-agent frameworks including ChatDev and GPTSwarm, AaaS-AN achieves notable performance gains across various foundation models. In comparision to ChatDev, one of its key advantages lies in its ability to significantly reduce token consumption. This efficiency comes from the strategy of AaaS-AN to eliminate redundant historical messages and superfluous "chat" rounds. The benefit is especially pronounced in code generation tasks, where large language models often struggle to pinpoint issues due to minor variations in lengthy and repetitive outputs.
To address this, AaaS-AN leverages a self-coordination mechanism that proactively initiates reflective reasoning when no substantive changes are detected in the generated code. This process effectively filters out unproductive code generation attempts, thereby enhancing overall performance while minimizing unnecessary token usage caused by ineffective agent actions. \par
% In comparison with state-of-the-art large language models, 多智能体协作能够有效解决代码生成过程中代码片段不完善、逻辑不严密的问题，并结合代码审查和测试过程与环境中进行多次交互从而提升较大。同时AaaS-AN能够有效地组织大模型性能较差的多智能体协作，显著降低花费成本。
In comparison with state-of-the-art large language models (as a single agent), multi-agent collaboration offers an effective solution to the common issues of incomplete code snippets and ill-structured logic in the code generation process.
By incorporating iterative interactions that integrate code review and testing within the execution environment, such collaboration significantly enhances the quality and reliability of the generated code. Furthermore, AaaS-AN is particularly effective at orchestrating multi-agent collaboration even when individual models demonstrate suboptimal performance. It achieves this while substantially reducing computational and token-related costs, thus improving overall efficiency and scalability in resource-constrained settings. \par
% AaaS-AN能够组织多智能体在智能体网络中自主协同以实现更好的性能，结构化的上下文也使得大模型能够从更短的提示词中准确感知智能体意图，从而减少token消耗。
AaaS-AN enables autonomous coordination among multiple agents within the agent network to achieve improved performance.
Its structured context allows large language models to more accurately infer agent intentions from shorter prompts, thereby reducing token consumption without sacrificing task effectiveness.
\begin{table}[htbp]
    \centering
    \fontsize{9}{12}\selectfont
    \begin{tabular}{ccccccc}
        \hline
        Benchmark & Method & Model & Quality & Token 
        & Cost & Time(s) \\
        \hline
        \multirow{6}{*}{\vspace{2.0cm}SRDD}
        & Qwen3-32b & - & 0.463 & 13707.569 & 0.056 &  128.606\\
        & GPT-4.1-Mini & - & 0.747 & 14294.313 & 0.086 & 125.707 \\
        & Claude-3.7-Sonnet & - & 0.584 & \underline{45442.749} &  \textbf{0.487} & 314.538 \\
        & \multirow{3}{*}{\vspace{0.8cm}GPTSwarm} & GPT-3.5-Turbo & - & - & - & - \\
        & & Qwen2.5-32b-Instruct & - & - & - & - \\
        & & Deepseek-V3& - & - & - & - \\
        & \multirow{3}{*}{\vspace{0.8cm}ChatDev} & GPT-3.5-Turbo & 0.839 & 21044.765 & 0.103 & 450.903 \\
        & & Qwen2.5-32b-Instruct & 0.891 & 29021.105 & 0.089 & 537.933 \\
        & & Deepseek-V3 & 0.854 & \textbf{59152.232} & \underline{0.202} & \textbf{1203.494} \\
        & \multirow{3}{*}{\vspace{0.8cm}AaaS-AN} & GPT-3.5-Turbo & 0.872 & 9037.320 & 0.044 & 279.534 \\
        & & Qwen2.5-32b-Instruct & \underline{0.899} & 11946.271 & 0.037 & 357.782 \\
        & & Deepseek-V3 & \textbf{0.900} & 23766.519 & 0.081 & \underline{887.330} \\
        \hline
        \multirow{6}{*}{\vspace{2.0cm}ProgramDev} 
         & Qwen3-32b & - & 0.452 & 15199.401 & 0.091 & 269.085 \\
        & GPT-4.1-Mini & - & 0.750 & 14411.400 & 0.121 & 108.444 \\
        & Claude-3.7-Sonnet & - & 0.838 & 21400.035 & \textbf{0.366} & 326.691 \\
        & \multirow{3}{*}{\vspace{0.8cm}GPTSwarm} & GPT-3.5-Turbo & - & - & - & - \\
        & & Qwen2.5-32b-Instruct & - & - & - & - \\
        & & Deepseek-V3& - & - & - & - \\
        & \multirow{3}{*}{\vspace{0.8cm}ChatDev} & GPT-3.5-Turbo & 0.767 & 24740.100 & 0.123 & 534.935 \\
        & & Qwen2.5-32b-Instruct & 0.716 & \underline{32189.967} & 0.095 & 555.670 \\
        & & Deepseek-V3 & 0.803 & \textbf{44078.333} & \underline{0.148} & \textbf{2145.833} \\
        & \multirow{3}{*}{\vspace{0.8cm}AaaS-AN} & GPT-3.5-Turbo & 0.774 & 10362.100 & 0.052 & 266.242 \\
        & & Qwen2.5-32b-Instruct & \textbf{0.900} & 14562.933 & 0.043 & 312.528 \\
        & & Deepseek-V3 & \underline{0.870} & 16990.633 & 0.057 & \underline{1163.419} \\
        \hline
    \end{tabular}
    \caption{The performance on the application-level code generation of SRDD and ProgramDev benchmark. The top scores are in bold, with the second-highest underlined in Quality.}
    \label{tab:code-generation}
\end{table}

\subsection{AaaS-AN Based Agent Services and Long-chain Flows}

%为进一步评估智能体网络解决通用任务的能力，我们收集了包含大量智能体服务与流程的数据集，并从任务、协议和服务三个角度对这些数据进行分析。
To further evaluate the capability of AaaS-AN in addressing general tasks, we collect a data set that includes a large number of intelligent agent services and processes and conduct an analysis of these data from the perspective of task, protocol, and service.

\subsubsection{Task Perspective}

\begin{table}
  \label{sample-table}
  \centering
  \begin{tabular}{ccccc}
    \toprule
    Task Status & Number & Average Length of Chain Flows & Average Time & Average Token Cost \\
    \midrule
    New & 919 (10.9\%) & 0.0 & 0.0 & 0.0 \\
    Running & 363 (4.3\%) & 3.6 & 57.1 & 1918.0 \\
    Success & 4518 (53.7\%) & 4.5 & 230.2 & 2254.6 \\
    Fail & 2620 (31.1\%) & 5.6 & 102.8 & 759.3 \\
    \bottomrule
  \end{tabular}
  \caption{Overview of Tasks}
\end{table}

\begin{table}
  \label{sample-table}
  \centering
  \begin{tabular}{cccccc}
    \toprule
    \multicolumn{6}{c}{Status} \\
    \cmidrule(r){2-6}
    Subtask & Total & New & Running & Success & Fail \\
    \midrule
    Agent & 44200 & 6870 (15.5\%) & 0 (0.0\%) & 28323 (64.1\%) & 9007 (20.4\%) \\
    RPA & 3055 & 4 (0.1\%) & 191 (6.3\%) & 2433 (79.6\%) & 427 (14.0\%) \\
    \bottomrule
  \end{tabular}
  \caption{Scale of Subtasks}
\end{table}

%我们的任务及子任务分为多种预定义的类型，包括新建(New)、运行中(Running)、成功(Success)和失败(Fail)。其中新建类型表示该任务仅仅只完成了最初的创建工作，运行中类型表示该任务处理挂起状态，可能是因为既没有完成也没有超时，成功和失败类型分别表示任务是否被顺利地完成。
Our tasks and subtasks are divided into several predefined types, including \textbf{New}, \textbf{Running}, \textbf{Success}, and \textbf{Fail}. The \textbf{New} type indicates that the task has only completed its initial creation. The \textbf{Running} type means that the task is in a pending state and has not been completed or timed out. The \textbf{Success} and \textbf{Fail} types, respectively, indicate whether the task was successfully completed or not. 

%从任务状态的分布来看，成功状态的任务数量最多，这表明 AaaS-AN 在大部分情况下能够顺利完成任务，体现出其稳定的优势。为保证稳定完成任务，需要更多的资源，因此成功状态的任务所消耗的时间和 Token 也更高。此外，失败的任务常常会执行错误的链路甚至是不断地循环，所以它们往往有着较长的链路。
From the perspective of task status distribution, the number of successful tasks reaches the highest level, indicating that AaaS-AN is capable of reliably accomplishing tasks in most cases, which reflects its stability. To ensure reliable task completion, more resources are typically required. Therefore, tasks in the successful status tend to consume more time and tokens.In contrast, \textbf{Fail} tasks often follow incorrect or even looping execution paths, resulting in a longer chain-flow. 

\subsubsection{Protocol Perspective}

\begin{table}
  \label{sample-table}
  \centering
  \begin{tabular}{ccccc}
    \toprule
    \multicolumn{5}{c}{Part} \\
    \cmidrule(r){2-5}
    Vertex & Number & Average Time & Average Token Cost & Success Rate \\
    \midrule
    Agent & 34947 & 19.8 & 710.8 & 97.1\% \\
    RPA & 2650 & 144.4 & - & 96.5\% \\
    \bottomrule
  \end{tabular}
  \caption{Protocol of Vertexes}
\end{table}

% AaaS-AN 具备良好的可扩展性，能够支持 Agents 和 RPAs等多种不同种类任务的类型，并且在处理这些任务时，有着出色的成功率。
AaaS-AN demonstrates good scalability, supporting various types of tasks such as those involving Agents and RPAs, while maintaining a high success rate in handling them.

\subsubsection{Service Perspective}

\graphicspath{{Figures/}}
\begin{figure}[htp]
    \centering
    \includegraphics[width=15cm]{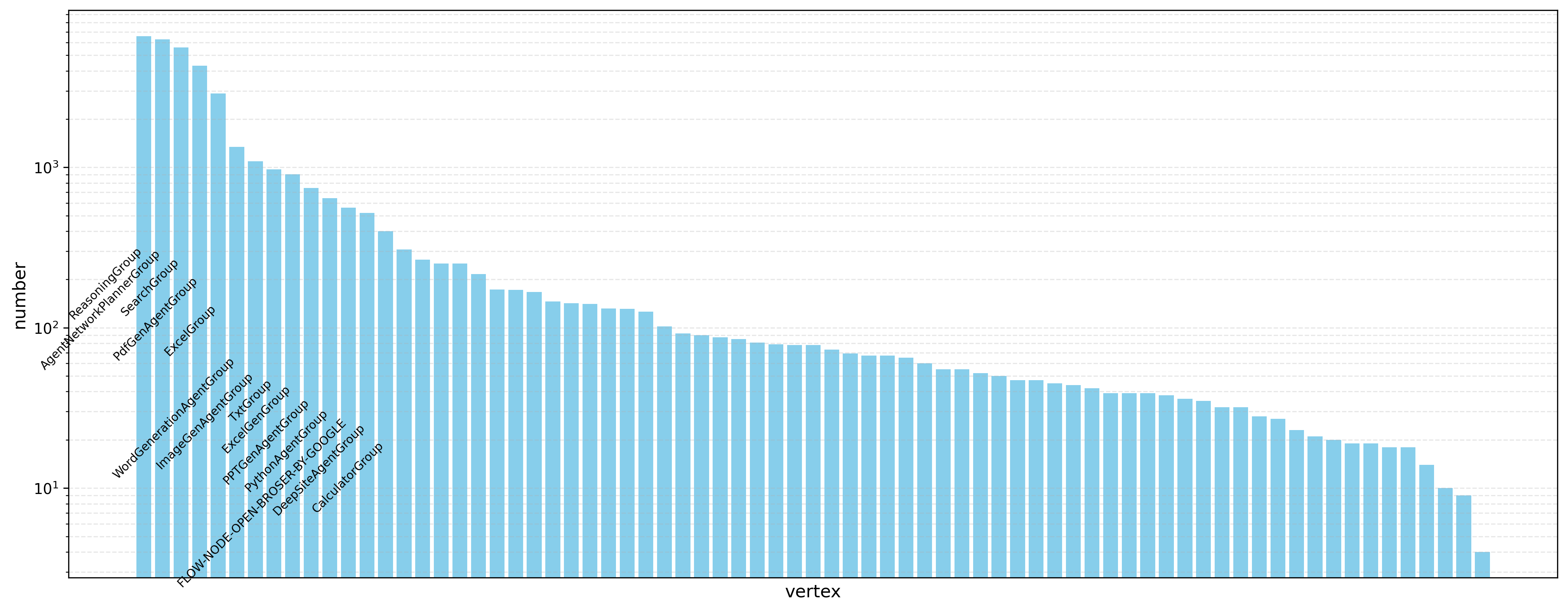}
    \caption{Distribution of Vertexes}
\end{figure}

%在 AaaS-AN 中，vertex是服务的基本单元。在处理任务数据时，我们发现各个 vertex 满足长尾分布特征，少数统计样本中不常见的 vertex 通常不会影响任务执行的效果，但是少数统计样本的种类繁多，也会累积一定的数量，因此这些统计样本在 AaaS-AN 中也有着很大的价值。vertex的分布如图2所示。
In AaaS-AN, vertexes are the fundamental units of service. During task data processing, it has been identified that the vertexes exhibit a long-tail distribution pattern. Although the uncommon vertices in limited statistical samples generally have minimal impact on task execution performance, their wide variety and accumulated volume grant them significant value within AaaS-AN. The distribution of vertexes is illustrated in Figure 2.

%TODO: 补充各个vertex在任务完成的贡献(contribution)

% AaaS-AN 调度的各个智能体服务(vertex)在完成任务的过程中发挥着不同的作用，我们将服务的贡献定义为服务的输入与任务最终输出的相似度的平均值，并收集了各个服务的成功率和贡献的数据，以发现较为优秀和需要进一步改进的服务，为 AaaS-AN 的进一步改进提供参考。
The agent services(vertexes) scheduled by AaaS-AN play different roles in the task execution process. We define the contribution of a service as the average similarity between its input and the final output of the task. We also collect statistics on the success rate and contribution of each service to identify those with strong performance and those requiring further improvement, thereby providing insights for future enhancements of AaaS-AN.

% \paragraph{Paragraphs}

\section{Conclusion}
We proposed AaaS-AN, a service-oriented framework for organizing large-scale multi-agent systems. Built upon the RGPS standard, AaaS-AN models agents and agent groups as dynamic network vertexes and integrates service discovery, registration, and execution through a unified scheduling mechanism. Our experiments on mathematical reasoning and application-level code generation demonstrate that AaaS-AN outperforms competitive baselines. We further validate its scalability via a deployment of over 100 agent services, including agent groups, RPA workflows and MCP servers. To facilitate future research, we release a dataset of over 10,000 multi-agent flows for evaluating long-chain tasks.

\bibliographystyle{unsrt}
\bibliography{neurips_2025}

\begin{thebibliography}{10}

\bibitem{DBLP:journals/corr/abs-2503-13415}
Weiqiang Jin, Hongyang Du, Biao Zhao, Xingwu Tian, Bohang Shi, and Guang Yang.
\newblock A comprehensive survey on multi-agent cooperative decision-making:
  Scenarios, approaches, challenges and perspectives.
\newblock {\em CoRR}, abs/2503.13415, 2025.

\bibitem{DBLP:journals/corr/abs-2402-01968}
Hung Du, Srikanth Thudumu, Rajesh Vasa, and Kon Mouzakis.
\newblock A survey on context-aware multi-agent systems: Techniques, challenges
  and future directions.
\newblock {\em CoRR}, abs/2402.01968, 2024.

\bibitem{DBLP:journals/corr/abs-2503-23278}
Xinyi Hou, Yanjie Zhao, Shenao Wang, and Haoyu Wang.
\newblock Model context protocol {(MCP):} landscape, security threats, and
  future research directions.
\newblock {\em CoRR}, abs/2503.23278, 2025.

\bibitem{hou2025modelcontextprotocolmcp}
Xinyi Hou, Yanjie Zhao, Shenao Wang, and Haoyu Wang.
\newblock Model context protocol (mcp): Landscape, security threats, and future
  research directions, 2025.

\bibitem{DBLP:books/sp/wsf14/WangFZHHZ14}
Jian Wang, Zaiwen Feng, Jia Zhang, Patrick C.~K. Hung, Keqing He, and
  Liang{-}Jie Zhang.
\newblock A unified rgps-based approach supporting service-oriented process
  customization.
\newblock In {\em Web Services Foundations}, pages 657--682. Springer, 2014.

\bibitem{gur2023real}
Izzeddin Gur, Hiroki Furuta, Austin Huang, Mustafa Safdari, Yutaka Matsuo,
  Douglas Eck, and Aleksandra Faust.
\newblock A real-world webagent with planning, long context understanding, and
  program synthesis.
\newblock {\em arXiv preprint arXiv:2307.12856}, 2023.

\bibitem{park2023generative}
Joon~Sung Park, Joseph O'Brien, Carrie~Jun Cai, Meredith~Ringel Morris, Percy
  Liang, and Michael~S Bernstein.
\newblock Generative agents: Interactive simulacra of human behavior.
\newblock In {\em Proceedings of the 36th annual acm symposium on user
  interface software and technology}, pages 1--22, 2023.

\bibitem{wang2024grutopia}
Hanqing Wang, Jiahe Chen, Wensi Huang, Qingwei Ben, Tai Wang, Boyu Mi, Tao
  Huang, Siheng Zhao, Yilun Chen, Sizhe Yang, et~al.
\newblock Grutopia: Dream general robots in a city at scale.
\newblock {\em arXiv preprint arXiv:2407.10943}, 2024.

\bibitem{liu2024caven}
Xiulong Liu, Sudipta Paul, Moitreya Chatterjee, and Anoop Cherian.
\newblock Caven: an embodied conversational agent for efficient audio-visual
  navigation in noisy environments.
\newblock In {\em Proceedings of the AAAI Conference on Artificial
  Intelligence}, volume~38, pages 3765--3773, 2024.

\bibitem{qian2023communicative}
Chen Qian, Xin Cong, Cheng Yang, Weize Chen, Yusheng Su, Juyuan Xu, Zhiyuan
  Liu, and Maosong Sun.
\newblock Communicative agents for software development.
\newblock {\em arXiv preprint arXiv:2307.07924}, 6(3), 2023.

\bibitem{chen2024internet}
Weize Chen, Ziming You, Ran Li, Yitong Guan, Chen Qian, Chenyang Zhao, Cheng
  Yang, Ruobing Xie, Zhiyuan Liu, and Maosong Sun.
\newblock Internet of agents: Weaving a web of heterogeneous agents for
  collaborative intelligence.
\newblock {\em arXiv preprint arXiv:2407.07061}, 2024.

\bibitem{wang2025user}
Lei Wang, Jingsen Zhang, Hao Yang, Zhi-Yuan Chen, Jiakai Tang, Zeyu Zhang,
  Xu~Chen, Yankai Lin, Hao Sun, Ruihua Song, et~al.
\newblock User behavior simulation with large language model-based agents.
\newblock {\em ACM Transactions on Information Systems}, 43(2):1--37, 2025.

\bibitem{dubois2023alpacafarm}
Yann Dubois, Chen~Xuechen Li, Rohan Taori, Tianyi Zhang, Ishaan Gulrajani,
  Jimmy Ba, Carlos Guestrin, Percy~S Liang, and Tatsunori~B Hashimoto.
\newblock Alpacafarm: A simulation framework for methods that learn from human
  feedback.
\newblock {\em Advances in Neural Information Processing Systems},
  36:30039--30069, 2023.

\bibitem{liu2024agents4plc}
Zihan Liu, Ruinan Zeng, Dongxia Wang, Gengyun Peng, Jingyi Wang, Qiang Liu,
  Peiyu Liu, and Wenhai Wang.
\newblock Agents4plc: Automating closed-loop plc code generation and
  verification in industrial control systems using llm-based agents.
\newblock {\em arXiv preprint arXiv:2410.14209}, 2024.

\bibitem{hong2023metagpt}
Sirui Hong, Xiawu Zheng, Jonathan Chen, Yuheng Cheng, Jinlin Wang, Ceyao Zhang,
  Zili Wang, Steven Ka~Shing Yau, Zijuan Lin, Liyang Zhou, et~al.
\newblock Metagpt: Meta programming for multi-agent collaborative framework.
\newblock {\em arXiv preprint arXiv:2308.00352}, 3(4):6, 2023.

\bibitem{wang2023plan}
Lei Wang, Wanyu Xu, Yihuai Lan, Zhiqiang Hu, Yunshi Lan, Roy Ka-Wei Lee, and
  Ee-Peng Lim.
\newblock Plan-and-solve prompting: Improving zero-shot chain-of-thought
  reasoning by large language models.
\newblock {\em arXiv preprint arXiv:2305.04091}, 2023.

\bibitem{wang2023unleashing}
Zhenhailong Wang, Shaoguang Mao, Wenshan Wu, Tao Ge, Furu Wei, and Heng Ji.
\newblock Unleashing the emergent cognitive synergy in large language models: A
  task-solving agent through multi-persona self-collaboration.
\newblock {\em arXiv preprint arXiv:2307.05300}, 2023.

\bibitem{qin2023toolllm}
Yujia Qin, Shihao Liang, Yining Ye, Kunlun Zhu, Lan Yan, Yaxi Lu, Yankai Lin,
  Xin Cong, Xiangru Tang, Bill Qian, et~al.
\newblock Toolllm: Facilitating large language models to master 16000+
  real-world apis.
\newblock {\em arXiv preprint arXiv:2307.16789}, 2023.

\bibitem{wei2022chain}
Jason Wei, Xuezhi Wang, Dale Schuurmans, Maarten Bosma, Fei Xia, Ed~Chi, Quoc~V
  Le, Denny Zhou, et~al.
\newblock Chain-of-thought prompting elicits reasoning in large language
  models.
\newblock {\em Advances in neural information processing systems},
  35:24824--24837, 2022.

\bibitem{wang2022self}
Xuezhi Wang, Jason Wei, Dale Schuurmans, Quoc Le, Ed~Chi, Sharan Narang,
  Aakanksha Chowdhery, and Denny Zhou.
\newblock Self-consistency improves chain of thought reasoning in language
  models.
\newblock {\em arXiv preprint arXiv:2203.11171}, 2022.

\bibitem{yao2023react}
Shunyu Yao, Jeffrey Zhao, Dian Yu, Nan Du, Izhak Shafran, Karthik Narasimhan,
  and Yuan Cao.
\newblock React: Synergizing reasoning and acting in language models.
\newblock In {\em International Conference on Learning Representations (ICLR)},
  2023.

\bibitem{shinn2023reflexion}
Noah Shinn, Federico Cassano, Ashwin Gopinath, Karthik Narasimhan, and Shunyu
  Yao.
\newblock Reflexion: Language agents with verbal reinforcement learning.
\newblock {\em Advances in Neural Information Processing Systems},
  36:8634--8652, 2023.

\bibitem{shen2023hugginggpt}
Yongliang Shen, Kaitao Song, Xu~Tan, Dongsheng Li, Weiming Lu, and Yueting
  Zhuang.
\newblock Hugginggpt: Solving ai tasks with chatgpt and its friends in hugging
  face.
\newblock {\em Advances in Neural Information Processing Systems},
  36:38154--38180, 2023.

\bibitem{selfself}
Standard Self-Reflection.
\newblock Self-contrast: Better reflection through inconsistent solving
  perspectives.

\bibitem{DBLP:conf/acl/QianLLCDL0CSCXL24}
Chen Qian, Wei Liu, Hongzhang Liu, Nuo Chen, Yufan Dang, Jiahao Li, Cheng Yang,
  Weize Chen, Yusheng Su, Xin Cong, Juyuan Xu, Dahai Li, Zhiyuan Liu, and
  Maosong Sun.
\newblock Chatdev: Communicative agents for software development.
\newblock In {\em Proceedings of the 62nd Annual Meeting of the Association for
  Computational Linguistics (Volume 1: Long Papers), {ACL} 2024, Bangkok,
  Thailand, August 11-16, 2024}, pages 15174--15186, 2024.

\bibitem{DBLP:conf/icml/ZhugeWKFKS24}
Mingchen Zhuge, Wenyi Wang, Louis Kirsch, Francesco Faccio, Dmitrii Khizbullin,
  and J{\"{u}}rgen Schmidhuber.
\newblock Gptswarm: Language agents as optimizable graphs.
\newblock In {\em Forty-first International Conference on Machine Learning,
  {ICML} 2024, Vienna, Austria, July 21-27, 2024}, 2024.

\end{thebibliography}

\end{document}